\documentclass{article}
\usepackage{spconf,amsmath,graphicx}

\usepackage{todonotes}
\usepackage{amsmath}
\usepackage{multirow}
\usepackage{latexsym}

\title{Echo State Speech Recognition}
\name{\hspace{-1.3ex}{\centering Harsh Shrivastava$^{1}$\sthanks{Work performed while interning at Google.}
    \hspace{0.2cm} 
    Ankush Garg$^{2}$
    \hspace{0.2cm}
    Yuan Cao$^{2}$
    \hspace{0.2cm}
    Yu Zhang$^{2}$
    \hspace{0.2cm}
    Tara Sainath$^{2}$
    \hspace{0.2cm}
    }}
\address{
    $^1$Georgia Institute of Technology\hspace{4ex} $^2$Google Inc.\\
    \texttt{\normalsize hshrivastava3@gatech.edu,\{ankugarg,yuancao,ngyuzh,tsainath\}@google.com}\\
    }
\begin{document}
%
\maketitle
\begin{abstract}
We propose automatic speech recognition (ASR) models inspired by echo state network (ESN) \cite{jaeger2001}, in which a subset of recurrent neural networks (RNN) layers in the models are randomly initialized and untrained. Our study focuses on RNN-T and Conformer models, and we show that model quality does not drop even when the decoder is fully randomized. Furthermore, such models can be trained more efficiently as the decoders do not require to be updated. By contrast, randomizing encoders hurts model quality, indicating that optimizing encoders and learn proper representations for acoustic inputs are more vital for speech recognition. Overall, we challenge the common practice of training ASR models for all components, and demonstrate that ESN-based models can perform equally well but enable more efficient training and storage than fully-trainable counterparts.
\end{abstract}
\section{Introduction}
\label{sec:intro}
Modern neural automatic speech recognition (ASR) models often contain tens or even hundreds of millions of parameters, and it is a conventional procedure to train every single model parameter through back-propagation. It has rarely been questioned whether such a heavy training procedure that aims to optimize every parameter is necessary at all. Answering this question not only helps us to better understand the training dynamics, providing insights into the sensitivity of each model component to the optimization procedure, but also potentially enable discovery of novel training procedures.

In this paper, we study this topic for ASR models based on recurrent neural networks (RNN). RNN has traditionally been an important building block in popular speech models \cite{graves2012sequence,graves2013speech,chiu2018state,rao2017explore,sainath2020streaming,gulati2020conformer} due to its excellent ability in capturing time-dependencies in sequential signals. We investigate whether training such models end-to-end is necessary at all to reach good performance. Our study is inspired by the formulation of echo state network (ESN) \cite{jaeger2001,maass2002}, which is a special type of RNN whose recurrent and input matrices are randomly generated and untrained. Despite this simple and counter-intuitive construction of RNN models, randomized recurrent connections demonstrated surprisingly good performance in capturing dynamics of a wide variety of time-series modeling tasks\cite{JAEGER2007335,Gallicchio2019ComparisonBD,Gallicchio2019a}. It is therefore an intriguing topic to study whether ESN can also work properly for modern ASR models.

Our study is focused on two types of models, namely RNN-T \cite{graves2012sequence,graves2013speech} and Conformer\cite{gulati2020conformer}. We experimented with replacing different trainable RNN components in these models with ESNs: for RNN-T we replaced either the encoder or prediction network, and for Conformer only the decoder is replaced. We conducted experiments on the Librispeech dataset \cite{panayotov2015libri} as well as long-form examples, and summarize our findings as follows:\\

\noindent\textbf{Randomized decoder performs equally well}: By replacing decoder RNN layers with ESN, the model quality remains almost the same as the fully trainable baselines. In fact, we even observed word-error-rate (WER) reduction on long-form examples with randomized decoders across multiple settings. This indicates that in ASR models, the dynamics of the decoder RNN is relatively simple and can be effectively absorbed even by randomly constructed networks.\\

\noindent\textbf{By contrast, randomized encoder hurts model quality}: We observed significant increase in WER when encoder RNN layers are randomized. Therefore in a fully trainable model, the encoder assumes the heavy-lifting and critical learning task of capturing meaningful representations for acoustic models.\\

\noindent\textbf{Randomized model can be trained and stored more efficiently}: Since ESN is randomly constructed and untrained, it does not go through back-propagation hence the training speed can be improved. We observed 37\% training speed gain for RNN-T models with ESN decoders. What is more, an ESN model can be deterministically regenerated from the same random seed used for building the network, we only need to store one random seed instead of the whole model when storage space is limited.

We give a brief introduction to ESN in Sec. \ref{sec:esn_intro}, and describe our proposed methodology in Sec. \ref{sec:esn_asr}. Our experimental results are reported in Sec. \ref{sec:experiments}, followed by related work and conclusion.


\section{Echo State Network}
\label{sec:esn_intro}
Echo State Network \cite{jaeger2001} is a special type of recurrent neural network, in which the recurrent matrix (known as ``reservoir'') and input transformation are randomly generated then fixed, and the only trainable component is the output layer (known as ``readout''). A very similar model named Liquid State Machine (LSM) \cite{maass2002} was independently proposed almost simultaneously, but with a stronger focus on computational neuroscience. This family of models started by ESN and LSM later became known as Reservoir Computing (RC) \cite{Verstraeten2007}.

A basic version of ESN has the following formulation :
\begin{align}
    \mathbf{h}_{t} &= \text{tanh} \left(\mathbf{W}_{res}\mathbf{h}_{t-1} + \mathbf{W}_{in} \mathbf{x}_t\right) \label{eq:esn} \\
    \mathbf{y}_t &= f(\mathbf{W}_{out}\mathbf{h}_{t}) \notag
\end{align}
in which $\mathbf{h}_t$ and $\mathbf{x}_t$ are the hidden state and input at time $t$,  $\mathbf{y}_t$ is the output and $f$ being a prediction function (for example softmax for classification). This formulation is almost equivalent to a simple RNN, except that the reservoir and input transformation matrices $\mathbf{W}_{res}$ and $\mathbf{W}_{in}$ are randomly generated and fixed. $\mathbf{W}_{res}$ is also often required to be a sparse matrix. The only component that remains to be trained is the readout weights $\mathbf{W}_{out}$. 

Despite the extremely simple construction process of ESN, it has been shown to perform surprisingly well in many regression and time-series prediction problems. A key condition for ESN to function properly is called the Echo State Property (ESP) \cite{jaeger2001,Yildiz2012}, which basically claims that the ESN states asymptotically depend only on the driving input signals (hence states are ``echos'' of inputs), while the influence of the initial states vanishes over time. ESP essentially requires the recurrent network to have a ``fading memory'', which is also shown to be critical in optimizing a dynamical system's computational capacity 
\cite{Legenstein2007}.

Theoretical analysis shows that in order for ESP to hold, the spectral radius of the reservoir matrix $\rho(\mathbf{W}_{res})$, defined as the largest absolute value of its eigenvalues, needs to be smaller than 1. 
Intuitively, $\rho(\mathbf{W}_{res})$ determines how long an input signal can be retained in memory: smaller radius results in a shorter memory while larger radius enables a longer memory. In addition, the scale of the input, which determines how strong inputs influence the dynamics, remains a hyperparameter critical to the performance of the model.

Recently ESNs have also been extended to deep versions in which multiple recurrent layers are stacked up \cite{Gallicchio2017,Gallicchio2017b},
It has been shown that different levels of the ESN layers are able to capture signal dynamics at different scales. 

\section{Echo State Speech Recognition Model} \label{sec:esn_asr}
\subsection{Model Architectures}
\begin{figure}[htbp]
\begin{minipage}[b]{.48\linewidth}
  \centerline{\includegraphics[width=4.5cm]{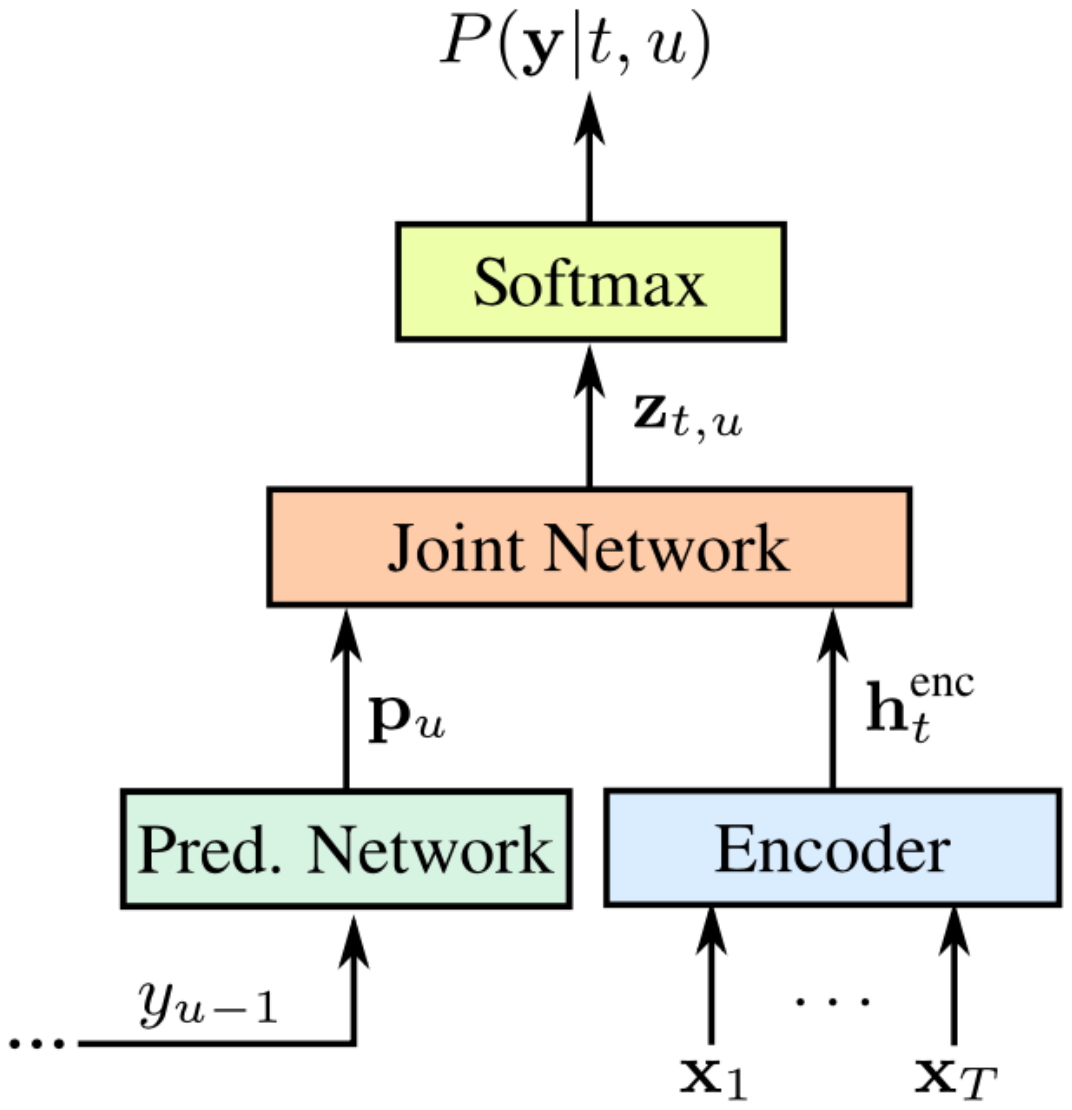}}
\end{minipage}
\begin{minipage}[b]{0.48\linewidth}
  \centerline{\includegraphics[width=4.5cm]{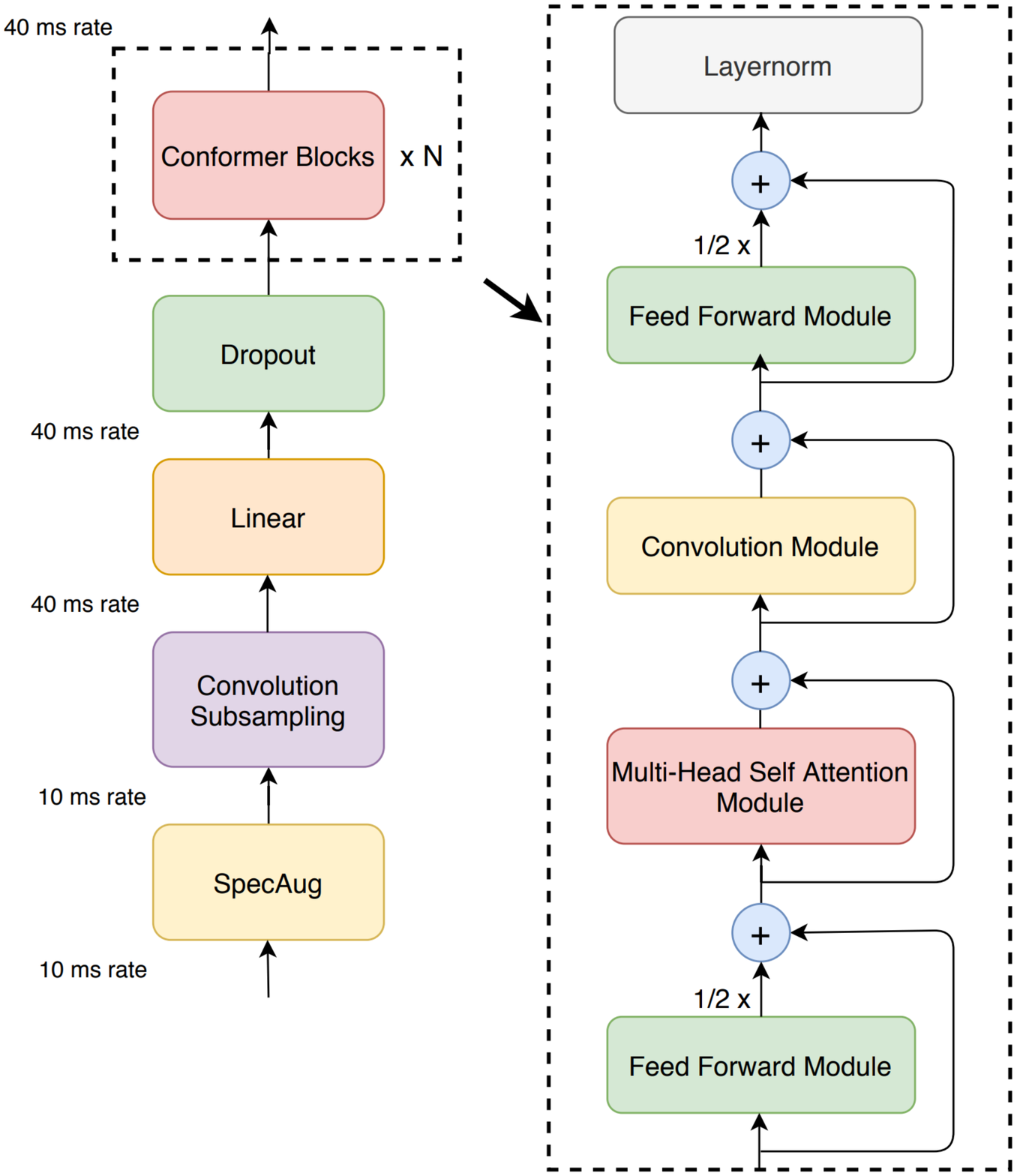}}
\end{minipage}
\caption{Architecture of RNN-T (left) and Conformer encoder (right).}
\label{fig:model_arch}
\end{figure}

Inspired by the intriguing property of ESN, we are interested in studying the behavior of ESN for ASR tasks. Our study is based on two backbone models: RNN-T \cite{graves2012sequence,graves2013speech} and Conformer \cite{gulati2020conformer}, two successful ASR model architecture that have achieved superior performances. The RNN-T model consists of an encoder as acoustic model, a prediction network (decoder), and a joint network. The major components of the encoder and decoder are RNN layers, usually using LSTM \cite{Hochreiter1997} as the recurrent cell. The Conformer model innovated the encoder with a mixture of convolutional layers and Transformer \cite{ashish2017} as building blocks, while using LSTM as decoder layers. The improved encoder enables more efficient representation learning for acoustic inputs, yielding state-of-the-art performace on Librispeech \cite{panayotov2015libri} benchmarks. The architectures of RNN-T and Conformer encoder are summarized in Fig.~\ref{fig:model_arch}

We propose to replace the RNN layers in both RNN-T and Conformer with ESN layers. For the RNN-T model, we replace either the encoder or decoder RNNs with ESNs (denoted by RNNT-E and RNNT-D respectively), but only decoder RNNs are replaced for the Conformer model (denoted by Conformer-D). All other model components remains trainable.

As described in Section \label{sec:esn_intro}, two critical hyperparameters that determine the dynamics of ESN and its behavior are the spectral norm of the reservoir matrix and input scale. While it is a common practice to tune these hyperparameters manually, we treat them as trainable scalars and let the optimization procedure find the suitable values. Specifically, we modify the ESN layer in Eq.~\ref{eq:esn} into

\begin{align}
    \mathbf{h}_t^l = \text{tanh} \left(\rho^l \mathbf{W}_{res}^{l}\mathbf{h}_{t-1} + \gamma^l \mathbf{W}_{in}^{l} \mathbf{x}_t\right) \label{eq:esn_new}
\end{align}

\noindent where $\rho^l$ and $\gamma^l$ are learnable scaling factors for the reservoir of the $l^{th}$ layer and input transformation matrices respectively.

Since our ESN adopts the simple RNN cell instead of LSTM as used by trainable RNN-T and Conformer decoder, the RNN layer parameters is 75\% less.

\subsection{Training} \label{sec:training}
In traditional ESN settings, usually the optimal parameter values for the readout layer $\mathbf{W}_{out}$ are obtained by solving the inverse problem. However, for more complex problems like ASR, it is not possible to use these inverse solvers as there are other trainable components involved. We therefore refer to back-propagation for training as in the base model case.

Note that since the recurrent layer weights are fixed and untrained, no gradient needs to be computed the common problem of gradient explosion and diminishing encountered can be alleviated. We hypothesize that such a light-weight training procedure not only speeds up the training procedure, but may also improve optimization efficiency, especially for long input sequences due to better-conditioned gradient flow.

\section{Experiments}
\label{sec:experiments}
\subsection{Data}
We conduct our experiments on Librispeech \cite{panayotov2015libri} which provides 970 hours of labelled speech along with 800M tokens text only corpus for language modeling. The models are evaluated on "test clean" and "test other" splits. Additionally, we also evaluate our models on long-form dataset, which is constructed by randomly concatenating utterances from the "test other" split. This yields 98 examples with minimum utterance length of 100 seconds and the maximum length of 350 seconds. We use this dataset to evaluate model performance on longer sequences.

\subsection{Models}
As mentioned in Sec.\ref{sec:esn_asr}, we explore two model types namely RNN-T and Conformer as our base architectures. RNN-T models contain 2 RNN layers in decoder network, 640 dimensional joint network and 16k word piece vocabulary processed from text corpus in Librispeech using BPE \cite{sennrich2016}. The two convolutional layers are followed by two RNN layers to form the encoder. The dimension of the decoder RNN cell is set to 256 or 512 in our experiments.

For Conformer models, we follow the large-size setup in \cite{gulati2020conformer} which consists of 17 encoder layers and 1 decoder layer, except that for the decoder RNN cell we use 256 and 512 dimension as in the RNN-T case. We also follow the same training setup as described in the original work (without using a language model).

For our proposed ESN models, the ESN cell recurrent weight matrices are initialized from uniform distribution between $-1$ and $1$\footnote{We also experimented with Gaussian distribution, but observed similar performance.} and fixed. The same is done for input transformation matrices. Weight matrices in ESNs are usually sparsely constructed, in our experiments we set the sparsity level to 80\%, namely only 20\% of the matrix entries are sampled from the uniform distribution while the remaining are set to zero.

\subsection{Results}
\subsubsection{Main results}
In our first experiment, we replaced all encoder or decoder RNN layers in RNN-T, and only the decoder in Conformer. The results are presented in Table~\ref{table:main}.

\begin{table}[htbp]
\small
\centering
\begin{tabular}{lllll}
\hline \textbf{Model} &\textbf{Dec dim} &\textbf{test clean} &\textbf{test other} & \textbf{longform} \\ \hline
\multirow{2}{*}{RNNT} &256 & 6.9 & 18.7 & 18.9 \\
                      &512 & 6.8 & 18.6 & 18.1\\
\multirow{2}{*}{Conformer} &256 & 2.1 & 4.8 & 5.7\\
                           &512 & 2.1 & 4.7 & 5.6\\
\hline
\multirow{2}{*}{RNNT-E} &256 & 35.6 & 60.2 & 61.7\\
                       &512 & 32.2 & 56.6 & 57.2\\
\multirow{2}{*}{RNNT-D} &256 & 6.6 & 18.3 & 18.9\\
                        &512 & 6.3 & 18.0 & 17.5\\
\hline
\multirow{2}{*}{Conformer-D} &256 & 2.0 &4.7 & 5.8 \\
                             &512 & 2.0 &4.6& 5.7 \\
\hline                             
\end{tabular}
\caption{\label{table:main} WER comparison for different models on Librispeech test sets. RNNT-E and RNNT-D denote RNNT model with encoder and decoder replaced by ESN respectively; Conformer-D means conformer model with decoder replaced by ESN.}
\end{table}

From Table~\ref{table:main}, we have the following observations:

\begin{enumerate}

    \item Replacing decoder RNN layers with ESN does not hurt performance: For both RNNT-D and Conformer-D, compared with the fully trainable baselines it can be seen that WER never increases, and in fact in many case ESN model is even better (for example for the 512-dimensional RNNT case, WERs of RNNT-D are lower than baselines by 0.5 in all case). We suspect that this is because due to the improved training efficiency achieved by the removal of weight updates in the decoder, as mentioned in Sec.~\ref{sec:training}. This observation indicates that the dynamics of the ASR decoder is relatively simple and can be absorbed by straightforward constructions like ESN.
    
    \item Replacing all encoder RNN layers, by contrast, hurts performance significantly. The contrast between RNNT-E and RNNT-D indicates that it is critical for an ASR model to learn proper representations for the acoustic signals in the encoder, and that the space spanned by the randomized ESN cells is not effective enough to capture the full complexity of acoustic inputs.
    
\end{enumerate}

\subsubsection{Progressive Training of Encoder Layers}
We further investigate the importance of training the RNN-T encoder by progressively making the encoder more trainable. The results are shown in Table~\ref{tbl:ablation_study}. We start with both encoder and decoder built with ESN layers, with 2 layers each. Keeping decoder random, we train one layer of encoder with trainable LSTM cell and observe that WER quickly drops to 8.9. We also observe that the choice of trainable layer, be it the first or second layer, doesn't impact model quality much. Making both the layers as trainable LSTM, WER further drops to 6.3. The trend suggests that both trainability and depth of the encoder is critical to ASR models.

\begin{table}[htbp]
\small
\centering
\begin{tabular}{cc}
\hline \textbf{Num. of ESN layers} &\textbf{WER} \\ \hline
2 & 34.5 \\
1 & 8.9 \\
0 & 6.3 \\
\hline
\end{tabular}
\caption{\label{tbl:ablation_study} 
Progressively training the encoder, keeping the decoder
fixed and random (ESN). Both encoder and decoder have 2 layers each and 512-dimension. First row corresponds to both encoder and decoder as random ESN. Last row corresponds to RNNT-D model in Table~\ref{table:main}}
\end{table}

\subsubsection{Training Speed and Storage Efficiency}
Since ESN layers require no weight update, gradients do not need to be computed for these layers and the models can be trained much faster. For example, in our experiments the 512-dimensional RNNT-D is 32\% faster than the trainable RNN-T (3.5 vs. 5.5 hours to reach 10k training steps)\footnote{The speed-up is not significant for Conformer-D as around 97\% of the Conformer model parameters are from the encoder, for which we cannot apply ESN.}. The speed-up can be potentially be more significant for both training and inference time if the hardware supports sparse matrix multiplication, since our ESN weight matrices are 80\% sparse.

On the other hand, since randomized ESN layers can be deterministically generated simply from one fixed random seed, to store the model offline we only need to save this single seed together with the remaining trainable model parameters. For example, in the fully trainable RNN-T model about 12\% (3411968 vs. 27980456) of the total model weights come from the decoder LSTM layers, which can be compressed into a single random seed in the case or RNNT-D, a significantly reduction in model size. This can be an appealing feature for on-device ASR models for which the installation package can be much smaller.




\section{Related Work}
\label{sec:related_work}
Although the concept of ESN and reservoir computing has been around for a long time, most of the applications are limited to time series analysis or signal processing. Their implication for speech recognition, especially in the deep learning age, has not been extensively studied. One early such investigation is \cite{Skowronski07}, in which they used ESN to predict the next frame speech features with discriminative training. \cite{Ghani10} use a simple recurrent neural reservoir for speech feature extraction which is then fed into a feedforward network for classification for each time step, on small-scale recognition tasks. 

Our findings that replacing decoders with randomized ESNs does not hurt model quality echos the results given by \cite{weinstein2020RNN}, in which they showed that RNN-T quality drops only slightly when the recurrent connections in the decoder layers are removed. Both studies indicate that the decoders do not model complex dynamics and can be a light-weight component, with our study verifying this from the perspective of randomized RNNs, which even outperformed trainable models in many cases, for both RNN-T and Conformer models.


\section{Conclusion}
\label{sec:conclusion}
In this paper, we investigated how a special type of RNN, namely echo state network whose recurrent and input weight matrices are purely randomly initialized and untrained, can be applied to ASR tasks. We proposed to replace a subset of RNN layers in RNN-T and Conformer models with ESN layers, and demonstrated that model quality does not drop or even perform better when the decoder is fully randomized. By contrast, randomizing encoders hurts model quality significantly, indicating that properly trained encoders are vital in learning proper representations for acoustic inputs. Our study challenges the common practice in which all ASR model components are fully trained, and showed that ESN-based models can perform equally well but admit much faster training speed.

\bibliographystyle{IEEEbib}
\bibliography{strings,refs}

\end{document}